\documentclass[11pt]{article}
\usepackage[T1]{fontenc}
\usepackage[letterpaper,margin=0.78in,top=0.40in,bottom=0.72in]{geometry}
\usepackage{times}
\usepackage[scaled=0.94]{helvet}
\usepackage{microtype}
\usepackage[table]{xcolor}
\usepackage{tcolorbox}
\tcbuselibrary{skins,breakable}
\usepackage{titlesec}
\usepackage{fancyhdr}
\usepackage{hyperref}
\usepackage{float}
\usepackage{fontawesome5}
\usepackage{placeins}

\definecolor{inkblue}{HTML}{173A5E}
\definecolor{accentblue}{HTML}{3E78A0}
\definecolor{frameblue}{HTML}{6B9EC1}
\definecolor{mistblue}{HTML}{F3F7FB}
\definecolor{ruleblue}{HTML}{173A5E}

\hypersetup{
  colorlinks=true,
  linkcolor=inkblue,
  citecolor=inkblue,
  urlcolor=accentblue
}

\titleformat{\section}
  {\sffamily\Large\bfseries\color{inkblue}}
  {\thesection}{0.75em}{}
\titleformat{\subsection}
  {\sffamily\large\bfseries\color{inkblue}}
  {\thesubsection}{0.75em}{}

\newtcolorbox{paperabstract}{
  enhanced,
  breakable,
  colback=mistblue,
  colframe=frameblue,
  boxrule=0.65pt,
  arc=3.2mm,
  left=25pt,
  right=25pt,
  top=14pt,
  bottom=16pt,
  boxsep=0pt,
  width=\textwidth,
  before skip=0pt,
  after skip=18pt
}

\makeatletter
\def\@papertitle{}
\def\@paperauthor{}
\newcommand{\papertitle}[1]{\gdef\@papertitle{#1}}
\newcommand{\paperauthor}[1]{\gdef\@paperauthor{#1}}

\newcommand{\makepaperfront}{%
  \thispagestyle{empty}%
  {\color{ruleblue}\rule{\textwidth}{2.5pt}}\par
  \vspace{1.05em}
  \begin{center}
    {\rmfamily\fontsize{21}{25}\selectfont\bfseries
      \@papertitle\par}
  \end{center}
  \vspace{0.65em}
  {\color{ruleblue}\rule{\textwidth}{1.5pt}}\par
  \vspace{1.45em}
  \ifx\@paperauthor\@empty
    \vspace{1.15em}% Reserved author area; intentionally blank for now.
  \else
    \begin{center}
      {\normalfont\rmfamily\small\color{black}\@paperauthor\par}
    \end{center}
  \fi
  \vspace{1.55em}
}

\newcommand{\paperresource}[2]{%
  \ifx#1\@empty
    \mbox{#2}%
  \else
    \href{#1}{\mbox{#2}}%
  \fi
}
\newcommand{\paperresources}{%
  \par\nobreak\vspace{14pt}%
  {\centering\rmfamily\bfseries\color{inkblue}\fontsize{11}{14}\selectfont
    \paperresource{\projectpageurl}{%
      \raisebox{-1.5pt}{{\fontsize{14}{16}\selectfont\faGlobe}}\hspace{0.5em}Project Page}%
    \hspace{2em}%
    \paperresource{\githuburl}{%
      \raisebox{-1.5pt}{{\fontsize{14}{16}\selectfont\faGithub}}\hspace{0.5em}Code}%
    \hspace{2em}%
    \paperresource{\huggingfaceurl}{%
      \raisebox{-3pt}{\includegraphics[height=17pt]{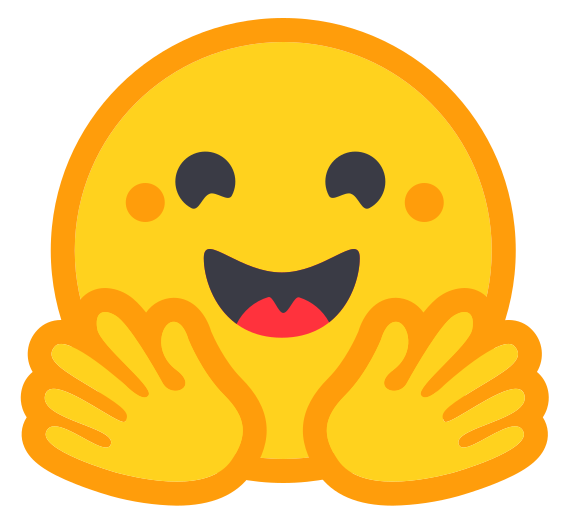}}\hspace{0.5em}Dataset}%
    \par}%
}
\makeatother

\usepackage{booktabs,array,makecell,graphicx}
\usepackage{xcolor,colortbl}
\usepackage{amsmath,amssymb,pifont}
\usepackage{wrapfig}

\providecommand{\memvenue}[1]{}
\renewcommand{\memvenue}[1]{%
  \nobreak\hspace{0.35em}{\normalfont\scriptsize\color{black!55}[#1]}}
\providecommand{\memyes}{}
\providecommand{\mempartial}{}
\providecommand{\memno}{}

\renewcommand{\memyes}{%
  \textcolor{green!45!black}{\ding{51}}}

\renewcommand{\mempartial}{%
  \textcolor{gray}{\ensuremath{\triangle}}}

\renewcommand{\memno}{%
  \textcolor{red!65!black}{\ding{55}}}

\papertitle{\textcolor[HTML]{245A81}{Mem}\textcolor[HTML]{238B8E}{In}\textcolor[HTML]{7355A6}{Stream}: Memorizing or Merely Recognizing?\\Rethinking Streaming Video Memory Evaluation}
\paperauthor{%
  Zhenqi He\textsuperscript{1}\quad
  Dunyuan Xu\textsuperscript{2}\quad
  Yang Bo\textsuperscript{3,}\footnote[2]{Project Leader.}\quad
  Kien T. Pham\textsuperscript{1}\quad
  Yanghao Wang\textsuperscript{1}\quad
  Shuhan Li\textsuperscript{3}\quad
  Xintong Guo\textsuperscript{3}\\[0.45em]
  Yuyang He\textsuperscript{3}\quad
  Fang Liu\textsuperscript{4}\quad
  Huankang Guan\textsuperscript{3}\quad
  Jingyan Chen\textsuperscript{3}\quad
  Rui Liu\textsuperscript{3}\quad
  Long Chen\textsuperscript{1,}\footnote[1]{Corresponding Author.}\\[0.75em]
  {\normalfont\small
    \textsuperscript{1}HKUST\quad
    \textsuperscript{2}CUHK\quad
    \textsuperscript{3}Huawei Research\quad
    \textsuperscript{4}CityU}\\[0.35em]
  {\normalfont\footnotesize zheci@connect.ust.hk,\quad longchen@ust.hk}
}

\newcommand*{\projectpageurl}{https://hkust-longgroup.github.io/MemInStream}
\newcommand*{\githuburl}{https://github.com/HKUST-LongGroup/MemInStream}
\newcommand*{\huggingfaceurl}{https://huggingface.co/datasets/zizi32/MemInStream}

\begin{document}

\begingroup
\renewcommand{\thefootnote}{\fnsymbol{footnote}}
\makepaperfront
\endgroup

\begin{paperabstract}
  \begin{center}
    {\rmfamily\fontsize{14}{17}\selectfont\bfseries\color{inkblue}Abstract}
  \end{center}
  \vspace{0.25em}
  Streaming video understanding aims to continuously process incoming visual observations, making the retention and retrieval of relevant information increasingly important as the video history grows. Although existing benchmarks include memory-related questions, question-answering accuracy alone does not reliably measure memory capabilities. Incorrect predictions may arise from failures in visual understanding rather than forgetting, whereas correct predictions may rely on linguistic priors or recent observations without requiring historical memory. To address these ambiguities, we introduce \textsc{MemInStream}, a benchmark dedicated to evaluating memory capabilities in streaming video understanding. We develop an evidence-controlled validation protocol that associates each question with temporally localized evidence intervals. The protocol validates evidence solvability and memory necessity by retaining questions that models can answer from the annotated evidence but cannot resolve using recent observations, post-evidence observations, or textual input alone. This helps disentangle memory limitations from understanding failures while reducing memory-independent shortcuts. \textsc{MemInStream} provides a systematic evaluation of memory capabilities through four major categories, covering Episodic Retrieval, State Tracking \& Updating, Multi-Evidence Integration, and Contextual Memory across diverse video scenarios. Overall, \textsc{MemInStream} provides a comprehensive and reliable benchmark for systematically evaluating memory capabilities in streaming video understanding.
  \paperresources
\end{paperabstract}

\section{Introduction}
\label{sec:introduction}
\begin{figure*}
    \centering
    \includegraphics[width=1\linewidth]{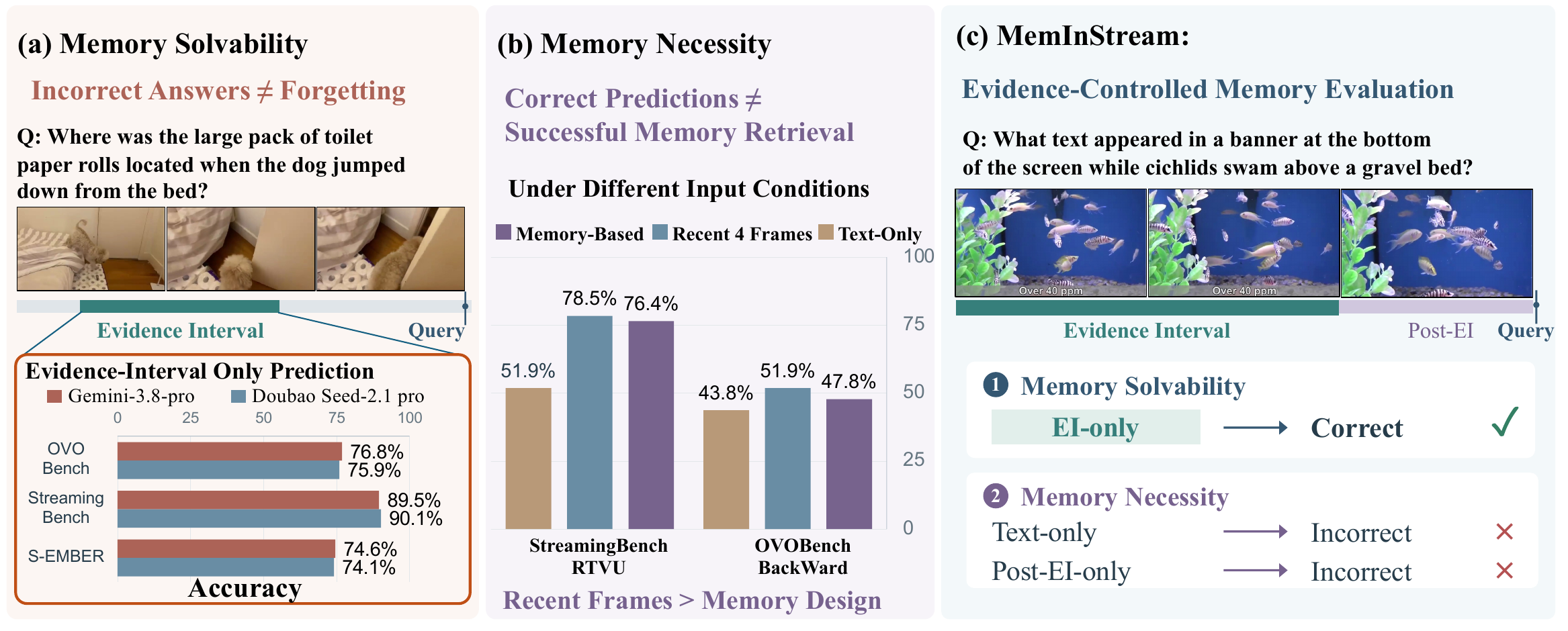}
    \vspace{-10pt}
    \caption{(a) \textbf{Memory Solvability}. Models may fail to answer questions even when the required evidence intervals are directly provided. Results are reported on the backward tracing tasks of OVO-Bench, the real-time video understanding tasks of StreamingBench, and S-EMBER. (b) \textbf{Memory Necessity}. We compare a memory-based method (FluxMem~\cite{xie2026fluxmem}) with SimpleStream~\cite{shen2026simple}, which uses only the four most recent frames, and a text-only baseline. (c) Our evidence-controlled validation protocol ensures both memory solvability and memory necessity, providing a comprehensive benchmark dedicated to evaluating memory capabilities in streaming video understanding.}
    \label{fig:teaser}
\end{figure*}
Streaming video understanding aims to continuously perceive and interpret incoming visual information, enabling models to track evolving events and respond to queries in real time~\cite{chen2024videollm,qian2025dispider,zhang2025flashvstream,yao2025timechatonline}.
Unlike offline video understanding, where the complete recording is available for analysis, streaming video understanding requires models to process observations sequentially as the video unfolds~\cite{zhou2024streamingdense}.
As visual history accumulates, models should retain relevant past information under limited memory and computational resources~\cite{xu2026streamingvlm,qian2024videostreaming}.
Such historical information remains essential for interpreting ongoing events and answering subsequent queries, even when the relevant observations are no longer available in recent frames~\cite{zeng2026onestreamer}.
This is particularly important for assistants expected to recall earlier activities and experiences~\cite{yang2025egolife}. For example, a user may ask where an object was previously placed or how its state has changed over time.
Effective streaming video understanding therefore requires not only understanding incoming observations, but also preserving, updating, and retrieving relevant information from the past, making memory a fundamental capability.

A growing number of benchmarks have been introduced to evaluate streaming video understanding~\cite{niu2025ovo,lin2024streamingbench,wang2026sember}.
Early efforts, such as StreamingBench~\cite{lin2024streamingbench}, assess real-time perception, omni-source understanding, and contextual understanding.
Subsequent benchmarks, including OVBench~\cite{huang2025online} and OVO-Bench~\cite{niu2025ovo}, further emphasize temporal awareness, incorporating memory retrieval and backward tracing alongside real-time perception and future-oriented reasoning. More recently, TeleEgo~\cite{yan2025teleego} and PhoStream~\cite{lu2026phostream} extend streaming evaluation to realistic egocentric and mobile scenarios, covering memory, cross-memory reasoning, and real-time audio-visual understanding. Dedicated memory benchmarks have also emerged: EgoStream~\cite{forte2026egostream} investigates episodic memory across different temporal horizons, S-EMBER~\cite{wang2026sember} focuses on temporally grounded episodic retrieval, and EgoProbe~\cite{pang2026egoprobe} evaluates long-range retention and event-specific recall. Despite these advances, performance on memory-related questions does not necessarily distinguish genuine memorization from mere recognition.

More specifically, two fundamental ambiguities arise when evaluating streaming memory through question-answering accuracy.
\textbf{(1) Unverified Memory Solvability.}
An incorrect answer does not necessarily indicate a memory failure, as models may struggle to perceive, interpret, or reason over the relevant visual information even when the necessary evidence is directly available. As shown in Fig.~\ref{fig:teaser}(a), recent video understanding models fail to answer nearly a quarter of the questions even when provided with the required evidence intervals.
Without verifying that models can answer a question when its supporting evidence is explicitly available, errors under streaming conditions cannot be reliably attributed to failures in retaining or retrieving historical information. \textbf{(2) Unverified Memory Necessity.}
Conversely, a correct answer does not necessarily demonstrate successful memory retrieval, as linguistic priors, answer-option cues, or recent visual observations may suffice without accessing historical evidence. As illustrated in Fig.~\ref{fig:teaser}(b), a text-only baseline achieves competitive accuracy on the evaluated benchmarks, while a method using only the four most recent frames~\cite{shen2026simple} even outperforms memory-based method~\cite{xie2026fluxmem}, which explicitly leverages historical memory. These results suggest that existing evaluations may permit shortcuts that bypass historical memory. Together, these observations highlight the need to establish not only whether historical evidence is sufficient to answer a question, but also whether it is necessary.

To address these ambiguities, we introduce \textsc{MemInStream}, a benchmark designed to systematically evaluate memory capabilities in streaming video understanding.
At its core, we develop an evidence-controlled validation protocol that assesses whether historical evidence is both sufficient and necessary to answer each question.
Specifically, each question is annotated with a query timestamp and evidence intervals containing the required historical information. As illustrated in Fig.~\ref{fig:teaser}(c), we first assess \textit{memory solvability} by directly providing the annotated evidence intervals to determine whether models can answer the question when the relevant information is available.
We then assess \textit{memory necessity} through text-only, recent-only, and post-evidence conditions, examining whether questions can be answered without the required historical evidence.
These complementary checks help exclude questions confounded by visual understanding difficulties or shortcuts that bypass historical memory. Furthermore, to evaluate diverse aspects of streaming memory, \textsc{MemInStream} comprises 17 fine-grained tasks organized into four categories: \textit{Episodic Retrieval} (recalling historical facts and events), \textit{State Tracking \& Updating} (maintaining and updating evolving states), \textit{Multi-Evidence Integration} (integrating information across historical observations), and \textit{Contextual Interactive Memory}.
Together, these designs enable a more systematic and targeted evaluation of memory capabilities in streaming video understanding.

Our contributions can be summarized as follows:
\textbf{(i) A comprehensive streaming video memory benchmark.}
We introduce \textsc{MemInStream}, a benchmark comprising four major memory categories and 17 fine-grained tasks across diverse scenarios, systematically assessing models' ability to retain, update, and retrieve historical information under streaming conditions.
\textbf{(ii) Evidence-controlled memory evaluation.}
We develop an evidence-controlled validation protocol that examines both the solvability and necessity of historical evidence, reducing the confounding effects of visual understanding limitations and memory-independent shortcuts in memory evaluation.
\textbf{(iii) Systematic evaluation of streaming memory capabilities.}
We evaluate representative streaming video models across diverse memory tasks and scenarios, examining their performance under different memory demands and highlighting areas for further improvement.

% Venue labels verified on 2026-10-20 against proceedings and official pages.
% StreamingBench: https://doi.org/10.1109/ICASSP55912.2026.11463959
% OVBench and OVO-Bench: CVPR 2025, official CVF proceedings.
% PhoStream: https://openreview.net/forum?id=cCMbxJcDmH (ICML 2026).
% TeleEgo, EgoStream, S-EMBER: retain arXiv years; conference venues unconfirmed.
\begin{table*}[t]
\centering
\caption{Comparison of streaming video benchmarks in terms of memory coverage and evidence-controlled validation. \memyes indicates explicit coverage, while \mempartial denotes partial coverage or related validation mechanisms.}
\label{tab:memory_benchmark_comparison}

\footnotesize
\setlength{\tabcolsep}{3pt}
\renewcommand{\arraystretch}{1.24}

\resizebox{\linewidth}{!}{%
\begin{tabular}{@{}l c c c c c c c c@{}}
\toprule
& \multicolumn{4}{c}{\textbf{Memory Coverage}}
& \multicolumn{2}{c}{\textbf{Evidence-Controlled Validation}}
& & \\
\cmidrule(lr){2-5}
\cmidrule(lr){6-7}

\textbf{Benchmark}
& \makecell{\textbf{Episodic}\\\textbf{Retrieval}}
& \makecell{\textbf{State Tracking}\\\textbf{\& Updating}}
& \makecell{\textbf{Multi-Evidence}\\\textbf{Integration}}
& \makecell{\textbf{Contextual}\\\textbf{Memory}}
& \makecell{\textbf{Evidence}\\\textbf{Solvability}}
& \makecell{\textbf{Evidence}\\\textbf{Necessity}}
& \makecell{\textbf{Evidence}\\\textbf{Interval}}
& \makecell{\textbf{Types of}\\\textbf{Videos}} \\
\midrule

StreamingBench\memvenue{ICASSP2026}
& \memyes & \mempartial & \mempartial & \memyes
& \memno & \mempartial\textsuperscript{a}
& \memno & General \\

OVBench\memvenue{CVPR2025}
& \memyes & \memyes & \memyes & \memno
& \memno & \memno
& \mempartial & General \\

OVO-Bench\memvenue{CVPR2025}
& \memyes & \mempartial & \memyes & \memno
& \memno & \memno
& \mempartial & Mixed \\

TeleEgo\memvenue{arXiv2025}
& \memyes & \memyes & \memyes & \memno
& \memno & \memno
& \memyes & Egocentric \\

PhoStream\memvenue{ICML2026}
& \memyes & \memno & \memyes & \mempartial
& \mempartial\textsuperscript{b} & \memno
& \memno & Mixed \\

EgoStream\memvenue{arXiv2026}
& \memyes & \memno & \memyes & \memno
& \memno & \mempartial\textsuperscript{c}
& \memyes & Egocentric \\

S-EMBER\memvenue{arXiv2026}
& \memyes & \mempartial & \memyes & \memno
& \mempartial\textsuperscript{d}
& \mempartial\textsuperscript{e}
& \memyes & Egocentric \\

\midrule
\rowcolor{blue!5}
\textbf{MemInStream}\memvenue{Ours}
& \memyes & \memyes & \memyes & \memyes
& \memyes & \memyes
& \memyes & Mixed \\
\bottomrule
\end{tabular}%
}

\par\vspace{3pt}
\begin{minipage}{\linewidth}
\scriptsize\raggedright
\textit{Validation notes:}
\mempartial\ denotes a related or partial control.
\textsuperscript{a}Humans remove questions answerable without video;
\textsuperscript{b}answers are verified using video prefixes;
\textsuperscript{c}text-only baseline evaluation;
\textsuperscript{d}ground-truth-guided sampling retains other history;
\textsuperscript{e}an LLM removes text-solvable questions.
\end{minipage}
\end{table*}

\section{Related Work}
\label{sec:related_work}

\subsection{Streaming Video Understanding}
\label{sec:related_streaming}

Streaming video understanding requires models to process continuously arriving observations and answer queries as events unfold, using only information available at the response time~\cite{chen2024videollm,qian2025dispider}. Unlike offline video understanding, which permits analysis of a complete recording, streaming systems must preserve relevant information before future queries are known. Memory is therefore essential for recalling past events, tracking changing states, and connecting temporally separated observations under bounded computational resources. Existing benchmarks examine these capabilities from complementary perspectives: StreamingBench~\cite{lin2024streamingbench} evaluates real-time visual, multisource, and contextual understanding. OVBench~\cite{huang2025online} and OVO-Bench~\cite{niu2025ovo} organize tasks around past, current, and future events. TeleEgo~\cite{yan2025teleego} extends evaluation to prolonged egocentric experiences and cross-memory reasoning, and PhoStream~\cite{lu2026phostream} assesses audiovisual understanding and response timing in mobile scenarios. However, QA accuracy can conflate memory with evidence understanding and memory-independent answer cues. A failure may reflect difficulty interpreting the relevant evidence even when it is available, while a correct answer may rely on recent observations rather than retained history. The competitive performance of SimpleStream using only recent frames underscores the need to verify whether historical memory is required~\cite{shen2026simple}.

Existing streaming memory methods differ primarily in how they retain and access historical information. HERMES~\cite{zhang2026hermes}, ReKV~\cite{di2025streaming}, and InfiniPot-V~\cite{kim2025infinipotv} reuse, retrieve, or compress KV caches, preserving previously computed attention states for subsequent queries. FluxMem~\cite{xie2026fluxmem}, VideoChat-Online~\cite{huang2025online}, and StreamForest~\cite{zeng2025streamforest} maintain compact visual representations through token compression or hierarchical memory banks, balancing historical coverage against storage and computation. Response-G1~\cite{ma2026responseg1} instead represents historical context through explicit scene graphs. Complementing these memory representations, training-based approaches such as StreamingVLM~\cite{xu2026streamingvlm}, AURA~\cite{lu2026aura}, JoyAI-VL-Interaction~\cite{joyai2026vlinteraction}, and MOSS-VL-Realtime~\cite{wang2026mossvl} adapt models to continuous observation and streaming responses through dedicated training. Despite differences in representation and training, these methods share the need to preserve useful information for future queries. MemInStream provides a common evaluation setting with evidence-controlled validation of memory solvability and necessity, distinguishing it from existing benchmarks, as summarized in Table~\ref{tab:memory_benchmark_comparison}.

\subsection{Video Memory}
\label{sec:related_video_memory}

Video memory benchmarks increasingly examine which historical details remain accessible, for how long, and under what temporal conditions. LongVideoBench~\cite{wu2024longvideobench} evaluates retrieval and reasoning over extended video and subtitle inputs, while S-EMBER~\cite{wang2026sember} grounds episodic recall in localized visual evidence. EgoStream~\cite{forte2026egostream} diagnoses recall across cognitive dimensions and temporal horizons using answer validity windows, and EgoProbe~\cite{pang2026egoprobe} tests both long-range retention and event-specific recall under interference from similar experiences. APM-Bench~\cite{huang2026apm} extends evaluation to persistent memory across interrupted sessions, while CapMem~\cite{liang2026capmem} examines textual captions as reusable episodic memory. These benchmarks broaden the temporal scope and specificity of memory evaluation, but answering accuracy still needs to be interpreted in relation to evidence availability and the model's ability to understand that evidence.

Video memory methods preserve historical information through compact visual representations or explicit, retrievable records. MovieChat~\cite{song2024moviechat} and MA-LMM~\cite{he2024mallm} compress and accumulate visual information in memory banks, extending access to past observations beyond the immediate input context. M3-Agent~\cite{long2025seeing} builds entity-centric episodic and semantic memories for subsequent retrieval and reasoning, while EgoLife~\cite{yang2025egolife} connects retrieval from extended egocentric recordings to everyday assistance through EgoRAG. Caption-based approaches studied in CapMem~\cite{liang2026capmem} provide an alternative by storing textual descriptions that can be reused across queries. Although these representations differ in storage cost and retained detail, assessing their effectiveness requires questions that both depend on historical evidence and remain answerable when that evidence is available. MemInStream makes these requirements explicit through joint memory solvability and necessity checks, providing evidence-controlled evaluation across four memory categories and 17 tasks.

\section{MemInStream Benchmark}
\label{sec:benchmark}

\subsection{Task Formulation}
\label{sec:task_formulation}
Streaming video memory evaluation assesses a model's ability to retain and utilize historical information while processing visual observations sequentially.
Let $V_{1:t}=(v_1,\ldots,v_t)$ denote the observations received up to time $t$.
The model incrementally updates its memory state and answers a question $q$ at query timestamp $t_q$:
\begin{equation}
    M_t=\mathcal U(M_{t-1},v_t), \qquad
    \hat a_q=\mathcal A(M_{t_q},q,\mathcal O_q),
\end{equation}
where $\mathcal U$ and $\mathcal A$ denote the memory update and answering functions, respectively, and $\mathcal O_q$ is the set of candidate answers.
The memory state $M_t$ is architecture-independent and may comprise cached visual tokens, compressed representations, textual summaries, or external memory.
Models must answer using their maintained memory, without unrestricted access to the original video history or future observations.

To identify the historical information required for each question, we annotate $K$ evidence intervals:
\begin{equation}
    \mathcal E_q=\{[s_i,e_i]\}_{i=1}^{K},
    \qquad 0\leq s_i<e_i<t_q,
\end{equation}
which localize the relevant visual evidence preceding the query.
For contextual memory tasks, we additionally annotate timestamped dialogue events $\mathcal H_q=\{(t_j,d_j)\}_{j=1}^{J}$, where $t_1<\cdots<t_J<t_q$, together with a subset $\mathcal R_q\subseteq\{1,\ldots,J\}$ identifying the necessary dialogue turns.
These dialogue events are incorporated into the model's memory sequentially, rather than provided as a complete transcript at query time.
Ground-truth answers, task labels, and evidence annotations are withheld from the streaming model and used only for evaluation.

\subsection{Taxonomy of Streaming Video Memory}
\label{sec:memory_taxonomy}

We organize streaming video memory into four categories and 17 fine-grained tasks, covering historical retrieval, state maintenance, temporal integration, and interaction-dependent memory. Tasks are defined by their primary memory requirements rather than question phrasing, and categories are not necessarily mutually exclusive.

\textbf{Episodic Retrieval.}
This category concerns retrieving information from specific past observations without continuous state tracking or integration across independent events. Its six tasks cover previously observed entities (\emph{Entity / Object Recall}), visual properties (\emph{Attribute Recall}), past actions or events (\emph{Action / Event Recall}), locations or spatial arrangements at a particular moment (\emph{Spatial / Location Recall}), event times or durations (\emph{Temporal Fact Recall}), and whether an entity or event was observed within a given period (\emph{Occurrence Verification}). 

\textbf{State Tracking \& Updating.}
This category requires maintaining and revising information as new observations arrive. \emph{Latest-State Retrieval} recovers an entity's most recently observed state; \emph{State Transition Tracking} follows its changes over time; \emph{Identity \& State Persistence} maintains entity identities and associated states across separated observations; and \emph{Correction / Revision Memory} updates information corrected or superseded by later evidence.

\textbf{Multi-Evidence Integration.}
This category combines historical observations when no single evidence interval suffices. \emph{Temporal Ordering} establishes chronological relationships among events; \emph{Sequential Action Integration} combines actions to infer an overall process or outcome; \emph{Counting} aggregates occurrences across time; \emph{Cross-Time Comparison} identifies similarities or differences between observations; and \emph{Multi-hop Memory Binding} connects information through intermediate entities, events, or relationships. 

\textbf{Contextual Memory.}
This category preserves information established through prior interactions for use in subsequent visual observations and queries. \emph{Entity--Identity Binding} retains user-defined associations between visual entities and names, identities, or roles. \emph{Conversational Context Memory} extends this requirement to references, constraints, and contextual information established across dialogue turns. Both require combining relevant conversational history with visual evidence, as in recalling Jerry's actions after an earlier identity assignment or resolving a reference to a previously specified object.
\section{Benchmark Construction}
\label{sec:construction}

\begin{figure}
    \centering
    \includegraphics[width=1\linewidth]{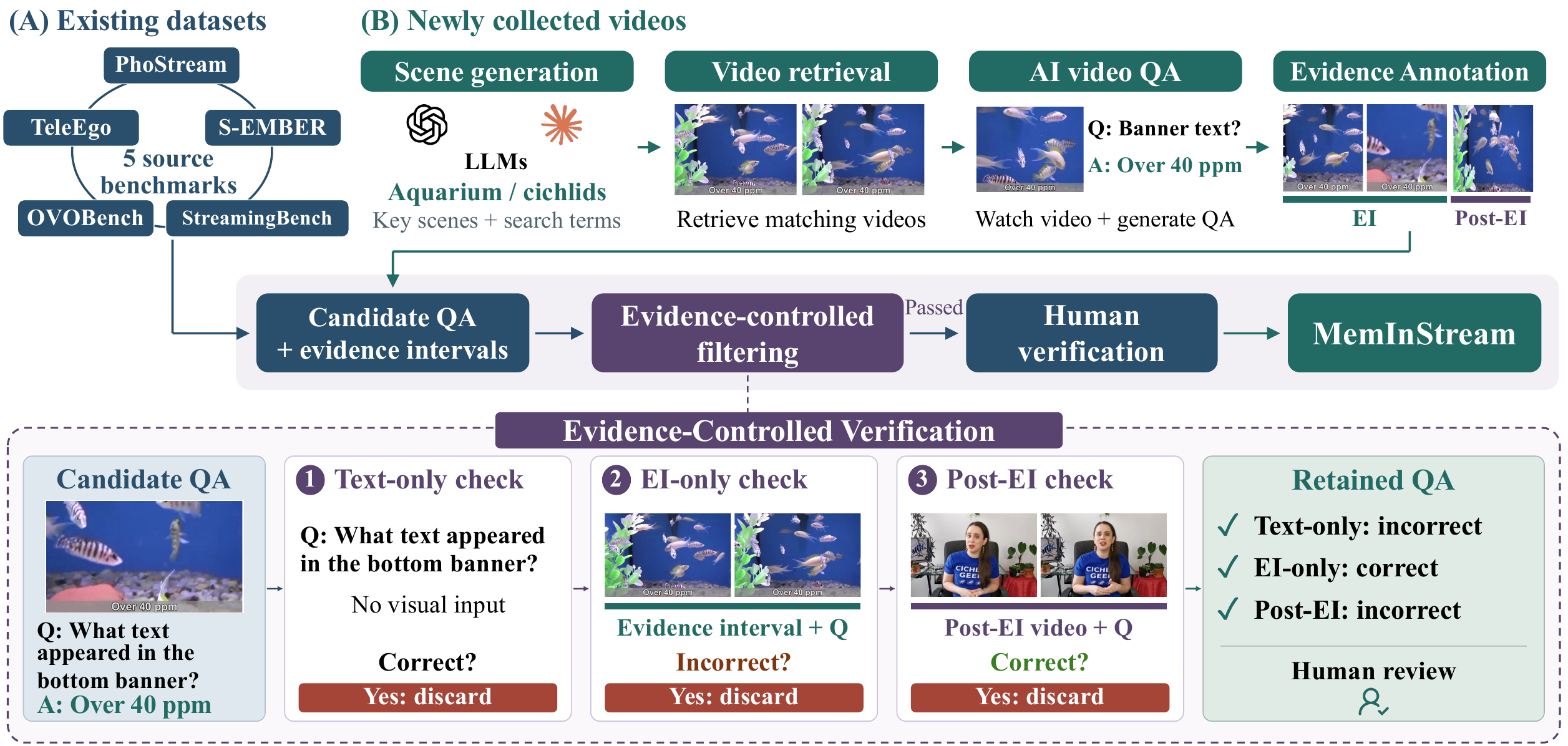}
    \caption{Overview of the \textsc{MemInStream} construction pipeline. 
    % Candidate questions are collected from five existing benchmarks or generated from newly retrieved videos. Each question is paired with historical evidence intervals and undergoes evidence-controlled validation to assess memory solvability and necessity, followed by human verification.
    }
    \label{fig:pipeline}
\end{figure}

In this section, we describe the construction of \textsc{MemInStream}, including candidate collection, question generation, evidence annotation, and quality validation. Our pipeline aims to provide comprehensive coverage of streaming memory capabilities while reducing the influence of understanding limitations and memory-independent shortcuts.

\subsection{Overall Pipeline}

As illustrated in Fig.~\ref{fig:pipeline}, we construct \textsc{MemInStream} from two complementary sources. First, we collect memory-related questions from existing streaming video benchmarks. Second, we retrieve additional videos and generate new questions to improve the coverage of memory capabilities and real-world scenarios. Candidates from both sources are standardized according to the task formulation in Sec.~\ref{sec:task_formulation}, with each question assigned a query timestamp and the historical evidence intervals required to answer it. All candidates subsequently undergo the same evidence-controlled validation principles, followed by human verification to obtain the final benchmark.

\subsection{Question and Evidence Annotation}

\textbf{Existing Benchmark Collection.}
We collect candidate questions from StreamingBench~\cite{lin2024streamingbench}, OVO-Bench~\cite{niu2025ovo}, TeleEgo~\cite{yan2025teleego}, PhoStream~\cite{lu2026phostream}, and S-EMBER~\cite{wang2026sember}. We focus on questions that require recalling or utilizing historical observations and organize them according to our unified memory taxonomy. Rather than directly incorporating these questions, we treat them as candidates for further evidence annotation and validation.

\textbf{New Video Collection and Question Generation.}
To complement existing benchmarks, we employ large language models to identify representative scenarios and generate corresponding video search terms for different memory tasks. We then retrieve relevant videos from the web and use video-language models to examine their content and generate candidate question-answer pairs. The newly generated questions target diverse memory requirements, including historical retrieval, state updating, multi-evidence integration, and contextual memory. All generated questions undergo the same subsequent annotation and validation procedures as those collected from existing benchmarks.

\textbf{Question and Answer Format.}
To balance objective evaluation with answer selection complexity, we adopt a unified format combining single-answer and multiple-answer questions. Each question contains a set of candidate options, with one or more correct answers. The number of correct options is not disclosed to the evaluated models, which are instructed to select all correct options. A prediction is considered correct only when the selected option set exactly matches the ground truth. 

\textbf{Evidence Annotation.}
For each candidate, we employ video-language models to examine the video and identify the historical observations necessary to answer the question. Specifically, we annotate the query timestamp $t_q$ and a set of evidence intervals $\mathcal E_q$ containing the required visual information. Depending on the memory task, the evidence may consist of a single historical episode or multiple temporally separated observations. For questions requiring multi-evidence integration, all relevant intervals must be identified to ensure that the annotated evidence jointly supports the answer.
For contextual memory questions, we additionally construct time-ordered dialogue histories and identify the historical turns necessary to resolve each query. These questions are designed to require both relevant dialogue context and visual evidence, such that neither source alone is sufficient to determine the answer. All annotations are subsequently examined during evidence-controlled validation and human verification.

\subsection{Evidence-Controlled Validation}
To disentangle memory from visual understanding and memory-independent shortcuts, we introduce an evidence-controlled validation protocol using two strong video understanding models, Gemini-3.8-Flash and Doubao Seed-2.1-Pro, to assess two complementary properties: \emph{memory solvability} and \emph{memory necessity}.

\noindent
\textbf{Memory Solvability.}
We assess whether a question can be answered when the required historical evidence is directly available. In the \emph{evidence-only} setting, both models receive the annotated evidence intervals, question, and answer options, without needing to retain information across the full video stream. We retain only questions correctly answered by both models, reducing the influence of visual understanding and reasoning difficulties.

\noindent
\textbf{Memory Necessity.}
We assess whether a question can be answered without the required historical evidence under two conditions.
The \emph{text-only} setting provides only the question and answer options to identify linguistic or option-level shortcuts.
The \emph{post-EI} setting provides the video segment from the end of the final evidence interval to the query timestamp, identifying answers recoverable from subsequent observations.
We retain only questions that both models answer incorrectly under each condition, reducing the likelihood that questions can be solved without historical memory.

\noindent
\textbf{Contextual Memory Validation.}
For contextual memory questions, we additionally verify their joint dependence on historical dialogue and visual evidence using the same two validation models.
We consider three conditions:
(1) \emph{Relevant dialogue + EI}, where both models must answer correctly given the necessary dialogue turns and evidence intervals;
(2) \emph{Irrelevant dialogue + EI}, where both models must answer incorrectly when the necessary dialogue turns are replaced with irrelevant historical turns;
and (3) \emph{Relevant dialogue + Post-EI}, where both models must answer incorrectly when the evidence intervals are replaced with post-EI observations.
We retain only questions satisfying all three conditions, reducing the likelihood that either historical dialogue or visual evidence can be bypassed.

\textbf{Automatic Filtering and Human Verification.}
During automatic validation, video frames are uniformly sampled at 4~fps. We employ Gemini-3.8-Flash and Doubao Seed-2.1-Pro as validation models, retaining only questions that satisfy the agreement-based solvability and necessity criteria described above. For questions with multiple correct options, correctness requires an exact match between the predicted and ground-truth option sets.
Following automatic filtering, human reviewers examine the retained questions against the original videos to verify answer correctness, temporal alignment of query timestamps and evidence intervals, and the relevance and completeness of the annotated evidence. Questions with ambiguous answers or inconsistent annotations are excluded.

% For contextual memory questions, the historical dialogue and necessary turns are additionally inspected. Reviewers assess answer correctness, evidence completeness, temporal consistency, dialogue dependence, and potential information leakage. Candidates with ambiguous questions, incorrect annotations, insufficient evidence, or alternative answer cues are corrected or discarded as appropriate. Together, these procedures reduce understanding-related confounds and memory-independent shortcuts, improving the reliability of streaming memory evaluation.

\begin{figure}[t]
    \centering
    \includegraphics[width=1\linewidth]{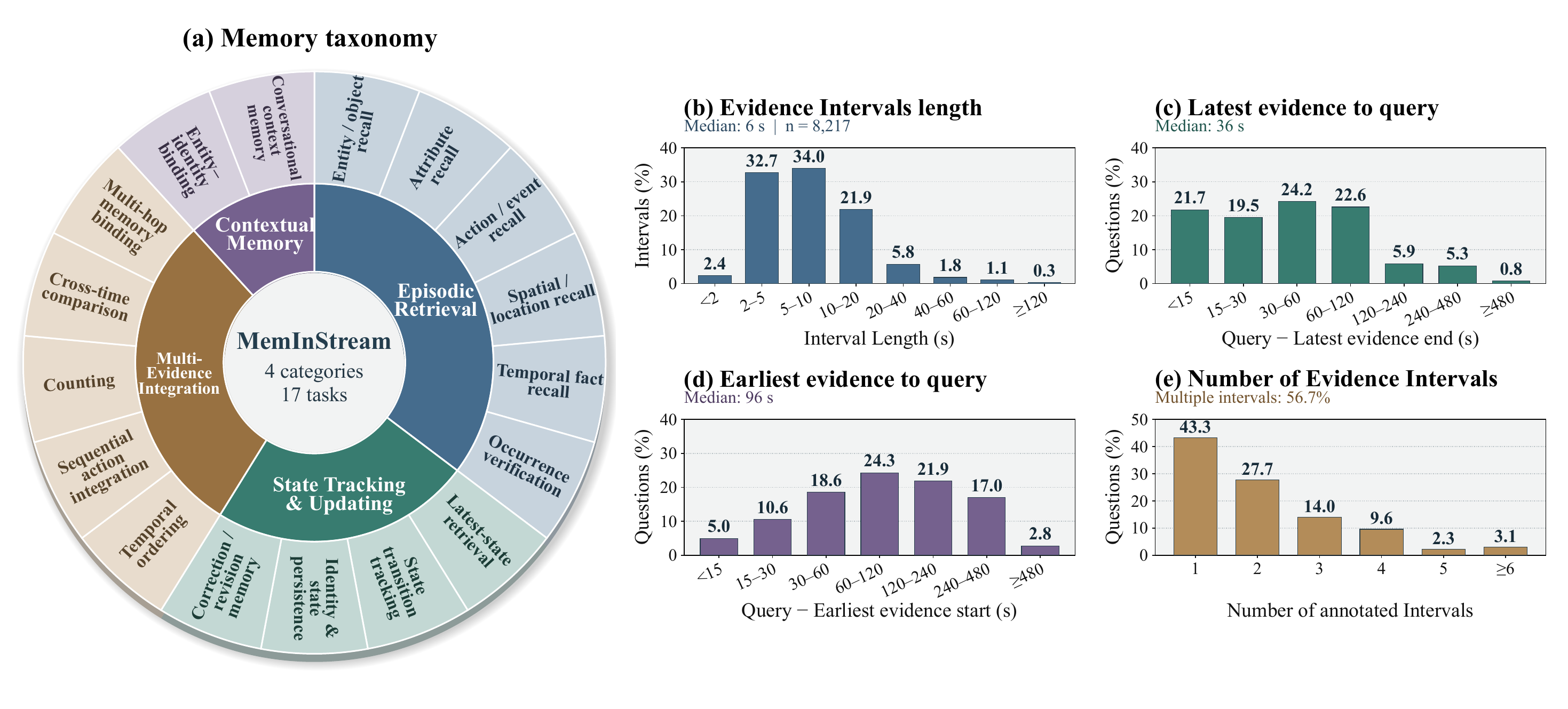}
    \vspace{-10pt}
    \caption{(a) Hierarchical taxonomy comprising four major memory categories and 17 fine-grained tasks. (b) Distribution of evidence interval lengths. (c--d) Temporal distances from the latest and earliest evidence intervals to the query, respectively. (e) Distribution of the number of evidence intervals per question.}
    \label{fig:stats}
\end{figure}
\subsection{Benchmark Statistics}

Fig.~\ref{fig:stats} provides an overview of the taxonomy and temporal characteristics of \textsc{MemInStream}. The benchmark contains 3,799 questions spanning four major memory categories and 17 fine-grained tasks, as shown in Fig.~\ref{fig:stats}(a). These questions are associated with 8,217 annotated evidence intervals, averaging 2.16 intervals per question. Each interval lasts 9.98 seconds on average, with a median duration of 6.0 seconds. Moreover, 56.7\% of questions involve multiple evidence intervals, highlighting the need to retain and integrate information from several historical segments.

To characterize the temporal demands of memory retrieval, we measure the distances from both the latest and earliest evidence to the query:
\[
d_q^{\mathrm{latest}} = t_q-\max_{[s,e]\in\mathcal E_q}e,
\qquad
d_q^{\mathrm{earliest}} = t_q-\min_{[s,e]\in\mathcal E_q}s.
\]
As illustrated in Fig.~\ref{fig:stats}(c--d), the median distances are 36 and 96 seconds, respectively. Notably, 34.6\% of questions have their latest evidence at least 60 seconds before the query, while 66.0\% require evidence beginning at least 60 seconds earlier, highlighting substantial temporal dependencies. Moreover, 56.7\% of questions involve multiple annotated evidence intervals, requiring models to retain and potentially integrate information across different moments. Together, these statistics demonstrate that \textsc{MemInStream} evaluates not only the retrieval of isolated historical observations but also the maintenance and integration of information over extended video streams.

\section{Experiment}
\subsection{Evaluation Protocol}
\label{sec:evaluation_protocol}
\textbf{Answer Format and Metrics.}
MemInStream adopts a unified evaluation format combining single-answer and multiple-answer questions, where models must select all correct options without knowing the number of correct answers.
Compared with conventional single-choice QA, this format requires models to identify the complete set of valid answers rather than selecting a single plausible option, reducing reliance on guessing.
Meanwhile, the structured answer space enables objective and reproducible evaluation, avoiding the ambiguity of assessing open-ended responses.
We adopt strict exact-match accuracy, where a prediction is correct only if the predicted answer set exactly matches the ground-truth set:
\begin{equation}
\mathrm{Acc} =
\frac{1}{N}
\sum_{i=1}^{N}
\mathbb{I}\left(
\hat{\mathcal{A}}_i = \mathcal{A}_i
\right),
\end{equation}
where $\hat{\mathcal{A}}_i$ and $\mathcal{A}_i$ denote the predicted and ground-truth answer sets, respectively.
We report overall accuracy and macro-averaged accuracy across memory tasks to account for differences in task sizes.

\begin{table*}[t]
\centering
\caption{Comparison on \textsc{MemInStream}. Task scores are exact-match accuracy (\%); each Avg. is the macro-average over tasks within that category. Frames/FPS reports input frame counts for proprietary and open-source offline models, and sampling rates (fps) for online models. Offline models use only the $N$ most recent sampled frames before the query.  $^{\dagger}$Response-G1 dynamically adjusts the fps between 0.25 and 1. $^{\ddagger}$Up to 60 recent frames per query. $^{\ast}$TimeChat-Online uses 4--316 input frames with adaptive sampling at 0.2, 0.5, or 1 fps. $^{\S}$Up to 768 input frames.}
\label{tab:meminstream_results}
\begingroup
\footnotesize
\setlength{\tabcolsep}{2.7pt}
\renewcommand{\arraystretch}{1.18}
\arrayrulecolor{black}
\resizebox{\linewidth}{!}{%
\begin{tabular}{@{}l c c !{\vrule width 0.4pt}
*{6}{c} !{\vrule width 0.4pt} c !{\vrule width 0.4pt}
*{4}{c} !{\vrule width 0.4pt} c !{\vrule width 0.4pt}
*{5}{c} !{\vrule width 0.4pt} c !{\vrule width 0.4pt}
*{2}{c} !{\vrule width 0.4pt} c @{}}
\toprule[0.8pt]
 & & & \multicolumn{21}{c}{{%
  \usefont{T1}{pzc}{m}{it}\fontsize{15}{18}\selectfont
  \textcolor[HTML]{245A81}{Mem}%
  \textcolor[HTML]{238B8E}{In}%
  \textcolor[HTML]{7355A6}{Stream}}} \\
\cmidrule(lr){4-24}
 & & & \multicolumn{7}{c}{Episodic Retrieval}
 & \multicolumn{5}{c}{\makecell{State Tracking\\\& Updating}}
 & \multicolumn{6}{c}{\makecell{Multi-Evidence\\Integration}}
 & \multicolumn{3}{c}{\makecell{Contextual\\Memory}} \\
\cmidrule(lr){4-10}\cmidrule(lr){11-15}\cmidrule(lr){16-21}\cmidrule(lr){22-24}
Method & Size & \makecell{Frames\\FPS}
 & EOR & AR & AER & SLR & TFR & OV & Avg.
 & LSR & STT & ISP & CRM & Avg.
 & TO & SAI & CNT & CTC & MMB & Avg.
 & EIB & CCM & Avg. \\
\midrule
\rowcolor{black!9}
\multicolumn{24}{c}{\textbf{\textit{Proprietary Models}}} \\
\midrule
Gemini-3.8-Flash & -- & 4 & 34.5 & 21.7 & 33.3 & 24.6 & 20.0 & 24.1 & 26.4 & 24.2 & 24.8 & 20.6 & 25.4 & 23.8 & 21.5 & 28.1 & 15.6 & 28.4 & 11.9 & 21.1 & 37.3 & 36.6 & 36.9 \\
Gemini-3.8-Flash & -- & 30 & 52.5 & 50.5 & 44.5 & 37.9 & 30.0 & 54.4 & 45.0 & 53.1 & 23.5 & 20.6 & 34.5 & 32.9 & 25.2 & 35.2 & 27.9 & 35.1 & 13.1 & 27.3 & 33.9 & 35.5 & 34.7 \\
Doubao Seed-2.1-pro & -- & 4 & 35.4 & 21.4 & 28.4 & 16.4 & 16.3 & 7.6 & 20.9 & 24.2 & 27.6 & 25.6 & 15.5 & 23.2 & 17.8 & 30.5 & 8.9 & 24.6 & 8.5 & 18.1 & 37.3 & 26.9 & 32.1 \\
Doubao Seed-2.1-pro & -- & 30 & 50.7 & 48.5 & 39.6 & 32.5 & 26.3 & 34.2 & 38.6 & 47.9 & 31.0 & 22.5 & 26.1 & 31.9 & 20.7 & 33.3 & 25.3 & 29.9 & 16.5 & 25.1 & 39.0 & 33.3 & 36.2 \\
\midrule
\rowcolor{black!9}
\multicolumn{24}{c}{\textbf{\textit{Open-Source Offline MLLMs}}} \\
\midrule
Qwen2.5-VL & 7B & 4 & 7.2 & 6.4 & 7.3 & 4.6 & 3.7 & 1.3 & 5.1 & 3.1 & 6.2 & 7.5 & 4.9 & 5.4 & 7.0 & 9.5 & 2.3 & 6.2 & 8.5 & 6.7 & 23.7 & 6.5 & 15.1 \\
Qwen2.5-VL & 7B & 30 & 9.4 & 10.7 & 5.7 & 4.6 & 2.2 & 10.1 & 7.1 & 4.6 & 5.5 & 6.9 & 4.9 & 5.5 & 4.1 & 11.4 & 1.3 & 6.2 & 6.3 & 5.8 & 22.0 & 7.5 & 14.8 \\
Qwen3-VL & 8B & 4 & 20.2 & 11.0 & 13.3 & 5.7 & 3.3 & 7.6 & 10.2 & 9.3 & 16.6 & 5.6 & 7.8 & 9.8 & 13.7 & 17.6 & 4.7 & 16.6 & 4.6 & 11.4 & 32.2 & 8.6 & 20.4 \\
Qwen3-VL & 8B & 30 & 25.6 & 29.1 & 16.4 & 11.4 & 9.4 & 12.7 & 17.4 & 18.0 & 11.0 & 5.6 & 11.3 & 11.5 & 10.0 & 19.1 & 7.6 & 16.6 & 2.8 & 11.2 & 37.3 & 2.2 & 19.7 \\
InternVL3.5 & 8B & 4 & 27.8 & 14.1 & 15.1 & 12.5 & 5.5 & 3.8 & 13.1 & 22.7 & 28.3 & 13.8 & 20.4 & 21.3 & 18.5 & 19.1 & 8.6 & 23.2 & 13.1 & 16.5 & 49.2 & 18.3 & 33.7 \\
InternVL3.5 & 8B & 30 & 29.2 & 34.8 & 16.2 & 12.1 & 6.5 & 19.0 & 19.6 & 31.4 & 19.3 & 13.8 & 18.3 & 20.7 & 17.4 & 20.5 & 11.2 & 20.9 & 8.5 & 15.7 & 39.0 & 16.1 & 27.6 \\
\midrule
\rowcolor{black!9}
\multicolumn{24}{c}{\textbf{\textit{Open-Source Online MLLMs (Training-Free)}}} \\
\midrule
FluxMem (Qwen2.5) & 7B & 1 & 11.7 & 16.1 & 6.5 & 5.7 & 2.9 & 13.9 & 9.5 & 2.6 & 2.8 & 3.1 & 6.3 & 3.7 & 1.9 & 6.7 & 1.0 & 5.7 & 3.4 & 3.7 & 17.0 & 4.3 & 10.6 \\
FluxMem (Qwen3) & 8B & 1 & 35.9 & 51.2 & 19.8 & 20.4 & 17.8 & 24.1 & 28.2 & 21.7 & 20.7 & 18.1 & 40.1 & 25.2 & 24.1 & 25.2 & 10.9 & 24.6 & 8.0 & 18.6 & 47.5 & 4.3 & 25.9 \\
Response-G1 (Qwen2.5) & 7B & 0.25--1$^{\dagger}$ & 6.7 & 7.4 & 5.0 & 3.6 & 4.1 & 6.3 & 5.5 & 2.1 & 3.5 & 1.3 & 8.5 & 3.8 & 4.8 & 4.8 & 2.6 & 4.3 & 2.8 & 3.9 & 6.8 & 0.0 & 3.4 \\
Response-G1 (Qwen3) & 8B & 0.25--1$^{\dagger}$ & 36.8 & 65.2 & 22.4 & 24.3 & 18.6 & 27.9 & 32.5 & 23.7 & 24.8 & 28.1 & 54.9 & 32.9 & 28.9 & 25.2 & 13.5 & 26.1 & 6.8 & 20.1 & 47.5 & 8.6 & 28.0 \\
HERMES (Qwen2.5) & 7B & 0.5 & 10.8 & 16.1 & 5.7 & 2.1 & 2.0 & 10.1 & 7.8 & 1.6 & 3.5 & 3.1 & 14.8 & 5.7 & 3.0 & 5.2 & 0.8 & 4.7 & 3.4 & 3.4 & 17.0 & 3.2 & 10.1 \\
\midrule
\rowcolor{black!9}
\multicolumn{24}{c}{\textbf{\textit{Open-Source Online MLLMs (Training-Based)}}} \\
\midrule
Flash-VStream (Qwen) & 7B & 1 & 30.9 & 31.1 & 28.4 & 20.4 & 30.2 & 17.7 & 26.5 & 34.0 & 21.4 & 29.4 & 59.2 & 36.0 & 39.6 & 30.5 & 21.4 & 36.0 & 16.5 & 28.8 & 50.9 & 19.4 & 35.1 \\
StreamForest  & 7B & 1 & 52.5 & 67.6 & 38.5 & 37.1 & 27.4 & 29.1 & 42.0 & 33.5 & 29.7 & 40.0 & 45.1 & 37.1 & 48.5 & 39.5 & 33.1 & 41.7 & 23.9 & 37.3 & 52.5 & 18.3 & 35.4 \\
TimeChat-Online & 7B & 0.2--1$^{\ast}$ & 17.9 & 36.8 & 7.6 & 8.6 & 12.2 & 19.0 & 17.0 & 7.2 & 14.5 & 6.3 & 54.2 & 20.5 & 20.0 & 21.0 & 18.0 & 16.1 & 5.7 & 16.1 & 37.3 & 0.0 & 18.6 \\
AURA & 8B & $2$ & 40.8 & 44.2 & 27.1 & 24.6 & 25.7 & 20.3 & 30.4 & 38.1 & 22.1 & 26.3 & 24.7 & 27.8 & 22.2 & 25.2 & 22.7 & 27.0 & 15.9 & 22.6 & 42.4 & 18.3 & 30.3 \\
JoyAI-VL-Interaction & 8B & $2$ & 53.4 & 65.6 & 32.0 & 32.1 & 31.2 & 31.6 & 41.0 & 43.3 & 35.9 & 28.8 & 67.6 & 43.9 & 39.6 & 37.1 & 30.7 & 37.4 & 18.2 & 32.6 & 50.8 & 8.6 & 29.7 \\
MOSS-VL-Realtime & 11B & $2^{\ddagger}$ & 26.9 & 31.1 & 14.8 & 11.8 & 11.6 & 12.7 & 18.2 & 18.6 & 17.9 & 10.6 & 9.9 & 14.2 & 9.6 & 9.5 & 14.6 & 16.6 & 4.0 & 10.9 & 18.6 & 9.7 & 14.2 \\
MOSS-VL-Realtime & 11B & $1^{\S}$ & 31.8 & 45.8 & 16.2 & 21.1 & 14.3 & 22.8 & 25.3 & 19.1 & 26.2 & 26.3 & 37.3 & 27.2 & 29.6 & 32.4 & 29.7 & 30.3 & 10.8 & 26.6 & 42.4 & 6.5 & 24.4 \\
\bottomrule[0.8pt]
\end{tabular}%
}
\endgroup
\par\vspace{3pt}
\begin{minipage}{\linewidth}
\scriptsize\raggedright
\textit{Task abbreviations.}
EOR: Entity / Object Recall; AR: Attribute Recall; AER: Action / Event Recall;
SLR: Spatial / Location Recall; TFR: Temporal Fact Recall; OV: Occurrence Verification;
LSR: Latest-State Retrieval; STT: State Transition Tracking; ISP: Identity \& State Persistence;
CRM: Correction / Revision Memory; TO: Temporal Ordering; SAI: Sequential Action Integration;
CNT: Counting; CTC: Cross-Time Comparison; MMB: Multi-hop Memory Binding;
EIB: Entity--Identity Binding; CCM: Conversational Context Memory.
\end{minipage}
\end{table*}

\subsection{Implementation Details}
\label{sec:implementation_details}

\textbf{Offline and proprietary models.}
We evaluate three open-source offline MLLMs, Qwen2.5-VL-7B~\cite{bai2025qwen25vl}, Qwen3-VL-8B~\cite{bai2025qwen3vl}, and InternVL3.5-8B~\cite{wang2025internvl35}, together with two proprietary models, Gemini-3.8-Flash~\cite{google2026gemini38flash} and Doubao Seed-2.1-pro~\cite{bytedance2026doubaoseed21pro}. Following the recent-frame protocol of SimpleStream~\cite{shen2026simple}, we evaluate these models with recent-frame budgets of $N\in\{4,30\}$ at 1 fps, using frames sampled from the $N$ seconds immediately preceding the query timestamp. Each model receives the available sampled frames in chronological order, together with the question and answer options, without maintaining persistent visual memory across queries.

\textbf{Online models.}
We divide open-source online MLLMs into training-free and training-based methods. The training-free group includes FluxMem~\cite{xie2026fluxmem}, Response-G1~\cite{ma2026responseg1}, and HERMES~\cite{zhang2026hermes}. FluxMem and Response-G1 are evaluated with both Qwen2.5-VL-7B and Qwen3-VL-8B backbones, while HERMES uses Qwen2.5-VL-7B. The training-based group comprises TimeChat-Online~\cite{yao2025timechatonline}, Flash-VStream~\cite{zhang2025flashvstream}, StreamForest~\cite{zeng2025streamforest}, AURA~\cite{lu2026aura}, JoyAI-VL-Interaction~\cite{joyai2026vlinteraction}, and MOSS-VL-Realtime~\cite{wang2026mossvl}. Online models process video observations chronologically and answer at the query timestamp using the context retained by their respective mechanisms.

\subsection{Main Results}
\label{sec:main_results}
\noindent
\textbf{Overall Performance.}
Tab.~\ref{tab:meminstream_results} presents the evaluation results across four memory categories and 17 fine-grained tasks.
Performance varies substantially across models and evaluation settings, highlighting the challenges of streaming video memory.
Among online models, StreamForest achieves the highest average accuracy across all four memory categories, with particularly strong performance on State Tracking \& Updating (37.06\%) and Multi-Evidence Integration (37.34\%), outperforming even the evaluated proprietary models in these categories.
Nevertheless, performance remains relatively limited across many tasks, suggesting that reliably retaining and utilizing historical information remains challenging for current streaming video models.

\noindent
\textbf{Challenges Across Memory Tasks.}
Different memory capabilities exhibit distinct performance characteristics.
In particular, Multi-hop Memory Binding remains challenging even for strong models, with Gemini-3.8-Flash (30 frames) and StreamForest achieving only 13.07\% and 23.86\%, respectively.
A similar difficulty emerges in Conversational Context Memory, where several online models perform substantially worse than on Entity--Identity Binding.
For example, StreamForest achieves 52.54\% on Entity--Identity Binding but only 18.28\% on Conversational Context Memory.
These results suggest that integrating temporally separated evidence and maintaining dialogue-dependent context present additional challenges beyond retrieving individual historical facts.

\noindent
\textbf{Impact of Visual Input and Model Design.}
Increasing the number of input frames generally improves episodic retrieval for offline models, with Gemini-3.8-Flash improving from 26.38\% to 44.96\% and Doubao Seed-2.1-Pro from 20.93\% to 38.63\% when increasing from 4 to 30 frames.
However, such improvements are not consistent across all memory categories, indicating that additional visual observations alone do not guarantee better memory performance.
Moreover, online methods exhibit considerable performance differences across backbone models.
For example, replacing Qwen2.5-VL with Qwen3-VL substantially improves both FluxMem and Response-G1 across all four categories.
These observations highlight the importance of both visual history coverage and underlying model capabilities when evaluating streaming memory methods.

\subsection{Analysis}
\label{sec:analysis}

\begin{figure*}[t]
    \centering
    \includegraphics[width=\linewidth]{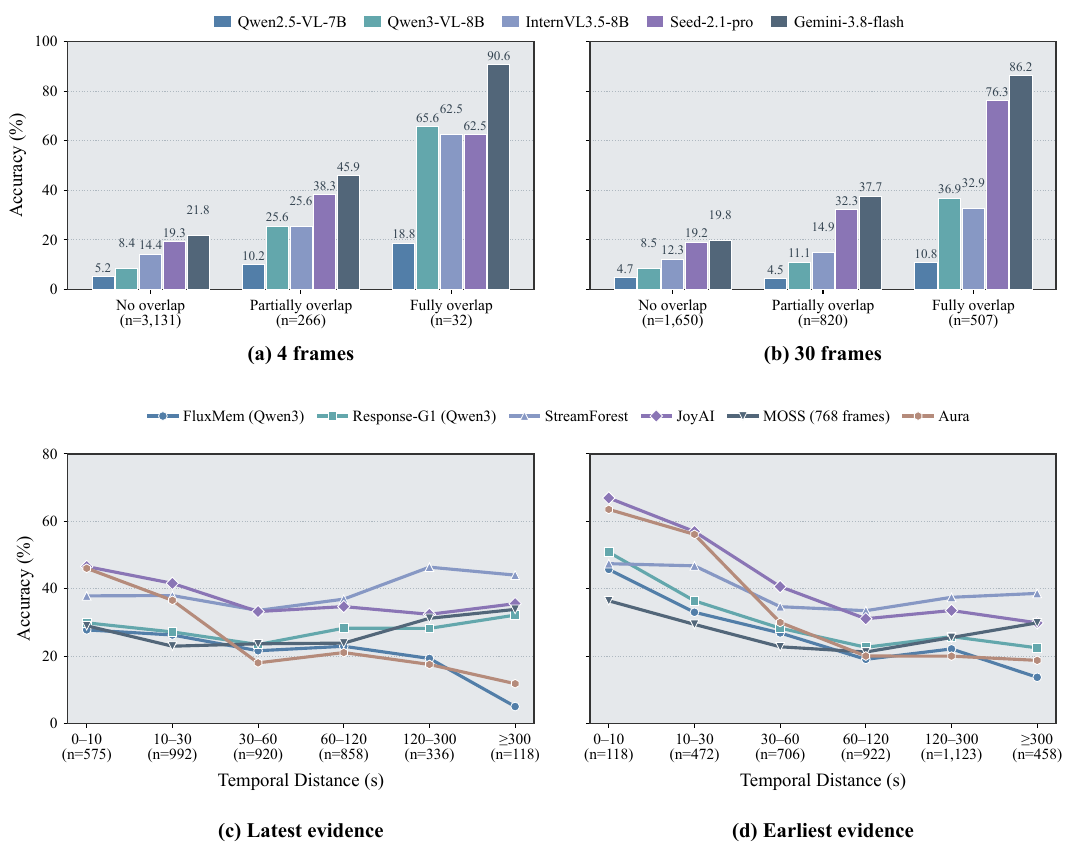}
     \vspace{-20pt}
    \caption{Accuracy under different evidence overlap levels and temporal distances. (a, b) Accuracy for 4-frame and 30-frame inputs, respectively. No overlap, partial overlap, and full overlap indicate that the input window covers none, some, or all of the annotated evidence intervals, respectively. (c, d) Accuracy versus temporal distance to the query, measured from the latest evidence end and the earliest evidence start, respectively. }
    \label{fig:evidence_overlap}
\end{figure*}

\noindent \textbf{Evidence Coverage and Memory Performance.}
Fig.~\ref{fig:evidence_overlap}(a-b) analyzes the relationship between recent-frame accuracy and the coverage of annotated evidence intervals. Across both frame budgets, all five models achieve substantially higher accuracy when the required evidence is fully covered by the input window. For instance, Gemini-3.8-Flash achieves 90.6\% and 86.2\% accuracy under full overlap with 4 and 30 frames, respectively, compared with only 21.8\% and 19.8\% without overlap. A similar trend is observed for Doubao Seed-2.1-pro, whose accuracy increases from 19.2\% to 76.3\% under the 30-frame setting. These results highlight the strong dependence of recent-frame performance on evidence availability. However, correct predictions without evidence overlap may arise from alternative visual cues, linguistic priors, or chance, rather than genuine historical memory. Conversely, even complete evidence coverage does not guarantee correct predictions, highlighting that access to the required evidence alone is insufficient and that effectively identifying and retrieving relevant information is equally important. 

\noindent\textbf{Temporal Distance and Memory Performance.}
Fig.~\ref{fig:evidence_overlap}(c-d) examines accuracy against the temporal distance from the latest evidence end and earliest evidence start to the query, respectively, reflecting the recency of relevant observations and the need to preserve earlier evidence.
AURA~\cite{lu2026aura}, which accesses only approximately 30 seconds of recent video, exhibits a sharp accuracy drop beyond this window: from 36.6\% to 18.0\% in (c) and from 56.1\% to 30.0\% in (d), between the 10--30s and 30--60s bins.
When the latest evidence ends more than 30 seconds before the query, all annotated evidence falls outside its accessible window.
In contrast, StreamForest~\cite{zeng2025streamforest}, Response-G1~\cite{ma2026responseg1}, and MOSS~\cite{wang2026mossvl} show less abrupt degradation, maintaining 44.1\%, 32.2\%, and 33.9\% accuracy, respectively, at distances of at least 300 seconds in (c).
JoyAI similarly achieves 35.6\% with memory enabled despite its limited recent-video window.
These results suggest the importance of preserving historical information beyond recent observations, although variations in question difficulty and composition across bins prevent interpreting these trends as isolated measures of memory decay.

% \begin{figure*}[t]
%     \centering
%     \includegraphics[width=\linewidth]{imgs/temporal_distance.pdf}
%     \caption{Accuracy versus temporal distance from (a) the latest evidence end and (b) the earliest evidence start to the query.}
%     \label{fig:temporal_distance}
% \end{figure*}

\begin{wrapfigure}{r}{0.5\textwidth}
    \vspace{-0.6\baselineskip}
    \centering
    \includegraphics[width=\linewidth]
        {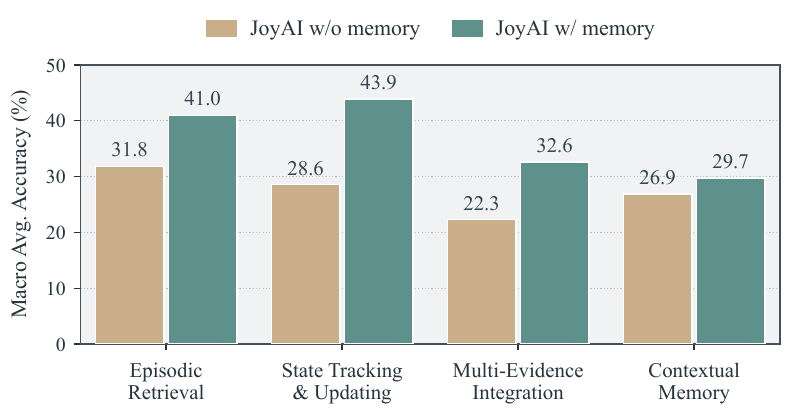}
    \vspace{-20pt}
    \caption{Effect of memory on JoyAI-VL-Interaction.}
    \label{fig:joyai_memory}
    
\end{wrapfigure}

\noindent\textbf{Sensitivity to Memory Augmentation.}
To examine whether \textsc{MemInStream} captures the benefit
of memory, we compare JoyAI-VL-Interaction~\cite{joyai2026vlinteraction} with and without
memory on the same evaluation questions, keeping the
recent-frame input budget fixed at up to 60 frames sampled
at 2 fps.
As shown in Fig.~\ref{fig:joyai_memory}, enabling memory
consistently improves macro-average accuracy across all
four Level 1 categories:
Episodic Retrieval increases from $31.8\%$ to $41.0\%$,
State Tracking \& Updating from $28.6\%$ to $43.9\%$,
Multi-Evidence Integration from $22.3\%$ to $32.6\%$,
and Contextual Memory from $26.9\%$ to $29.7\%$.
The largest gains occur in state tracking and
multi-evidence integration, suggesting that these tasks
particularly benefit from retaining and combining
historical information.
These results provide evidence that \textsc{MemInStream}
is sensitive to the utility of memory under a fixed
recent-observation budget.
This finding is consistent with our evidence-controlled
construction, which aims to select questions that require
historical evidence and cannot be reliably answered from
recent observations alone.
\FloatBarrier
\section{Conclusion}
\label{sec:conclusion}

We introduce \textsc{MemInStream}, a comprehensive benchmark for evaluating memory capabilities in streaming video understanding. To distinguish genuine memory performance from visual understanding limitations and memory-independent shortcuts, we develop an evidence-controlled validation protocol that examines both memory solvability and necessity. The resulting benchmark comprises 3,799 questions, spanning four memory categories and 17 fine-grained tasks. Extensive evaluations reveal substantial challenges for existing streaming video models, particularly in integrating historical evidence and maintaining contextual information. Further analysis shows that access to relevant historical evidence plays an important role in answering memory-related questions. By providing diverse memory tasks with evidence-controlled validation, \textsc{MemInStream} offers a systematic foundation for evaluating and advancing memory capabilities in streaming video understanding.

\noindent\textbf{Limitations and Future Work.}
While MemInStream evaluates memory capabilities across different temporal distances, its current scope does not fully capture the challenges of prolonged video streams. Future work will extend the benchmark to longer videos with greater temporal separation between evidence and queries, enabling more comprehensive evaluation of long-term memory retention, updating, and integration.

\bibliographystyle{unsrt}
\bibliography{references}

\end{document}